\documentclass{article}
\usepackage{ijcai19}

\usepackage{times}
\usepackage{soul}
\usepackage{url}
\usepackage[hidelinks]{hyperref}
\usepackage[utf8]{inputenc}
\usepackage[small]{caption}
\usepackage{graphicx}
\usepackage{amsmath}
\usepackage{booktabs}
\usepackage{algorithm}
\usepackage{algorithmic}
\usepackage{amssymb}
\usepackage{bm}
\usepackage{xcolor}
\title{Convolutional Auto-encoding of Sentence Topics \\for Image Paragraph Generation\footnote{This work was performed at JD AI Research.}}

\author{
Jing Wang$^1$\and
Yingwei Pan$^2$\and
Ting Yao$^{2}$\and
Jinhui Tang$^1$\footnote{Corresponding Author.}\And
Tao Mei$^2$\\
\affiliations
$^1$School of Computer Science and Engineering, Nanjing University of Science and Technology, China\\
$^2$JD AI Research, Beijing, China\\
\emails
\{jwang, jinhuitang\}@njust.edu.cn,
\{panyw.ustc, tingyao.ustc\}@gmail.com,
tmei@live.com
}
\begin{document}

\maketitle

\begin{abstract}
  Image paragraph generation is the task of producing a coherent story (usually a paragraph) that describes the visual content of an image. The problem nevertheless is not trivial especially when there are multiple descriptive and diverse gists to be considered for paragraph generation, which often happens in real images. A valid question is how to encapsulate such gists/topics that are worthy of mention from an image, and then describe the image from one topic to another but holistically with a coherent structure. In this paper, we present a new design --- Convolutional Auto-Encoding (CAE) that purely employs convolutional and deconvolutional auto-encoding framework for topic modeling on the region-level features of an image. Furthermore, we propose an architecture, namely CAE plus Long Short-Term Memory (dubbed as CAE-LSTM), that novelly integrates the learnt topics in support of paragraph generation. Technically, CAE-LSTM capitalizes on a two-level LSTM-based paragraph generation framework with attention mechanism. The paragraph-level LSTM captures the inter-sentence dependency in a paragraph, while sentence-level LSTM is to generate one sentence which is conditioned on each learnt topic. Extensive experiments are conducted on Stanford image paragraph dataset, and superior results are reported when comparing to state-of-the-art approaches. More remarkably, CAE-LSTM increases CIDEr performance from 20.93\% to 25.15\%.
\end{abstract}

\section{Introduction}
The recent advances in Convolutional Neural Networks (CNN) have successfully pushed the limits and improved the state-of-the-art technologies of image understanding. As such, it has become achievable to recognize an image with a pre-defined set of categories. In a further step to describe an image in a natural-language utterance, image captioning \cite{vinyals2015show,xu2015show,Rennie2016Self} has expanded the understanding from individual labels to a sentence to reflect the visual content in the image. Nevertheless, considering that a single sentence certainly has a upper-bound of descriptive capacity and thus fails to recapitulate every details in the image, the task of image paragraph generation is recently introduced in \cite{krause2017hierarchical,liang2017recurrent}. The ultimate goal is to generate a coherent paragraph that describes an image in a finer manner.

\begin{figure}[tb]
\centering
\includegraphics[width=0.46\textwidth]{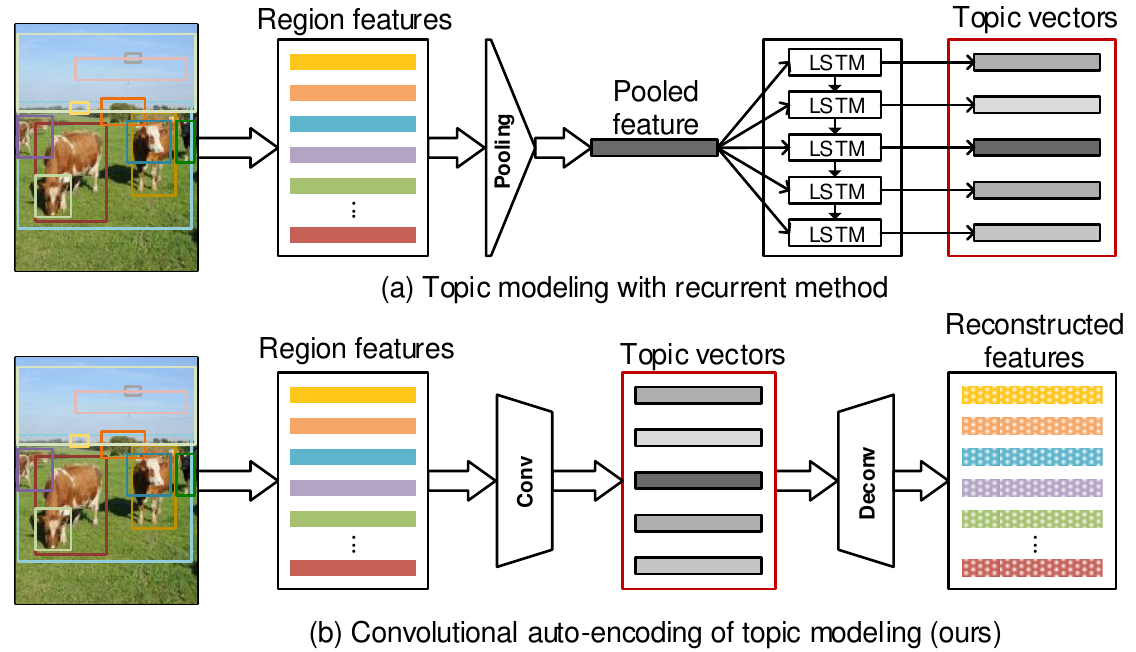}
\caption{Modeling the topics from an image by employing (a) recurrent method and (b) convolutional auto-encoding (ours).}
\label{fig:1}
\end{figure}

The difficulty of image paragraph generation originates from two aspects: 1) how to distill the gists/topics that are worthy of mention from an image? 2) how to describe each topic in one sentence while ensuring the coherence across sentences to form a paragraph? One straightforward way in \cite{krause2017hierarchical} to alleviate the first issue relies on Recurrent Neural Networks (RNN), e.g., Long Short-Term Memory (LSTM) network. The region-level features are encoded into a global vector via mean pooling, which is subsequently input into LSTM to decode the topics, as shown in Figure \ref{fig:1}(a). In this case, the inherent structure among all regions is not fully exploited, making it difficult to encapsulate the topics in the image. Furthermore, there is no clear evidence that the outputs of LSTM could characterize the topics well. Instead, we novelly devise Convolutional Auto-Encoding (CAE) structure to model the topics of an image in this paper as illustrated in Figure \ref{fig:1}(b). CAE, on one hand, abstracts the topics in the encoder by employing convolutions over the region-level features, and on the other, steers the deconvolutional decoder through reconstruction from topics to features. As such, the learnt topics are potentially more representative and contain the information needed. To address the second issue, we remould a two-level LSTM-based paragraph generation framework, in which paragraph-level LSTM models the dependency holistically across all the sentences in a paragraph and sentence-level LSTM generates words in sequence conditioning on each learnt topic.

\begin{figure*}[t]
\centering
\includegraphics[width=0.78\textwidth]{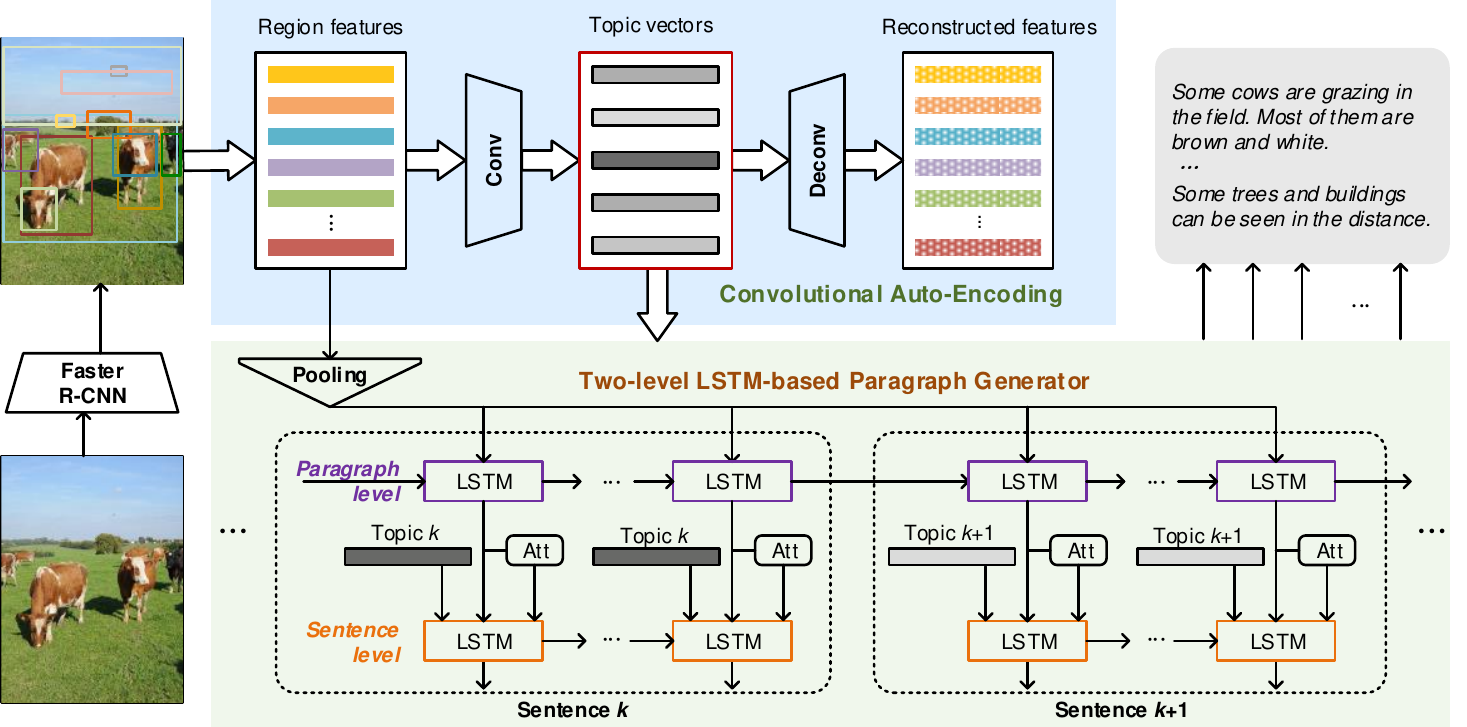}
\caption{An overview of our Convolutional Auto-Encoding plus Long Short-Term Memory (CAE-LSTM) architecture for image paragraph generation. All region-level features of image regions detected by Faster R-CNN are firstly injected into Convolutional Auto-Encoding (CAE) module for triggering the modeling of topics in the image. In particular, a convolutional encoder is leveraged to distill the knowledge from region-level features and extract the topics. Next, a deconvolutional decoder is steered to reproduce the region-level features from the distilled topics. As such, the learnt topics are ensured to capture the holistic and representative information that are worthy of mention from all regions. After that, we fuse all the region-level features via mean pooling as the image feature, which is fed into a two-level LSTM-based paragraph generator with attention (Att) for paragraph generation. Here the paragraph-level LSTM typically explores the inter-sentence dependency in a paragraph with the recursive input of mean-pooled image feature and the sentence-level LSTM generates the sentence conditioning on each learnt topic. Moreover, we include a reward of Coverage to encourage global coverage of objects in the paragraph for self-critical training.}
\label{fig:2}
\end{figure*}

By consolidating the idea of modeling sentence topics for image paragraph generation, we present a new Convolutional Auto-Encoding plus Long Short-Term Memory (CAE-LSTM) architecture, as shown in Figure \ref{fig:2}. Specifically, Faster R-CNN is firstly implemented to detect a set of salient image regions. We purely perform convolutions over all the region-level features in the encoder to distill the knowledge and extract the topics in the image. The learnt topics are ensured to capture holistic and representative information through achieving high reconstruction quality by the deconvolutional decoder. After that, we fuse all the region-level features via mean pooling as image representation, which are fed into a two-level LSTM networks for paragraph generation. The paragraph-level LSTM typically explores the dependency recursively throughout the procedure of paragraph generation and outputs a new initial state at a time for sentence-level LSTM with the paragraph history. As such, the generation of one sentence is affected by semantic context from the previous sentences when producing the paragraph. The sentence-level LSTM generates the sentence conditioning on each learnt topic, one word at each time step. Please also note that we uniquely design the metric of Coverage which is to encourage global coverage of objects in the sentence and consider such metric as a reward in the self-critical training strategy. The whole CAE-LSTM framework is jointly optimized.

The main contribution of this work is the proposal of modeling gists/topics in an image to boost image paragraph generation. This also leads to the elegant views of how to distill the knowledge and abstract the topics in an image, and how to nicely integrate the learnt topics into sentence generation and ensure inter-sentence dependency in a paragraph, which are problems not yet fully understood in the literature.

\section{Related Work}
\paragraph{Image Captioning.} The dominant paradigm in modern image captioning is sequence learning methods \cite{Donahue14,li2019pointing,vinyals2015show,xu2015show,You:CVPR16,Liu:2016PGSPIDEr,wang2018show,yao2017boosting,yao2018exploring,yao2017deep,Rennie2016Self} which utilize CNN plus RNN model to generate free-form and content-relevant captions. \cite{vinyals2015show} proposes an end-to-end neural network architecture by utilizing LSTM to generate sentence for an image. Later, by amending the discrepancy between training and inference distributions for sequence modeling, \cite{Rennie2016Self} presents a self-critical sequence training strategy to further enhance image captioning. Most recently, a combined bottom-up and top-down attention mechanism \cite{anderson2017bottom} is proposed for image captioning, which enables attention at the level of objects and salient image regions.

\paragraph{Image Paragraph Generation.} The task of image paragraph generation has received increasing attention recently, which describes an image with a long, descriptive, and coherent paragraph. \cite{krause2017hierarchical} is one of the early works that leverages a hierarchical RNN for generating paragraph with the input of region features. In particular,~a sentence RNN recursively generates sentence topic vectors conditioned on the global vector (i.e., the mean pooling of region-level features) and a word RNN is subsequently adopted to decode each topic into output sentence. \cite{liang2017recurrent} extends the hierarchical RNN by involving multi-level adversarial discriminators for paragraph generation. The paragraph generator is thus enforced to produce realistic paragraphs with smooth logical transition between sentence topics. Furthermore, \cite{chatterjee2018diverse} augments the hierarchical RNN with coherence vectors, global topic vectors, and a formulation of Variational Auto-Encoders \cite{kingma2013auto} to further model the inherent ambiguity of associating paragraphs with images.

\paragraph{Summary.} In short, our approach focuses on the latter scenario, which produces a coherent paragraph to depict an image. Unlike most of the aforementioned methods that recursively decode gists/topics for paragraph generation via LSTM conditioned on the global vector, our work contributes by exploiting a convolutional and deconvolutional auto-encoding module for gists/topics modeling over region-level features. Such design not only abstracts the topics in the convolutional encoder by exploiting inherent structure among all regions, but also steers the deconvolutional decoder through reconstruction to enforce the distilled topics more representative and informative. Moreover, a reward of Coverage is adopted in the self-critical training strategy to encourage global coverage of objects in the paragraph.

\section{Approach}
We devise our Convolutional Auto-Encoding plus Long Short-Term Memory (CAE-LSTM) architecture to facilitate image paragraph generation by modeling gists/topics in an image. CAE-LSTM firstly utilizes a convolutional encoder to encapsulate region-level features into the topics, which are endowed with the holistic and representative information through achieving high reconstruction quality by deconvolutional decoder. The distilled topics are further integrated into a two-level LSTM-based paragraph generator, enabling the inter-sentence dependency modeling in a paragraph via paragraph-level LSTM and topic-oriented sentence generation through sentence-level LSTM. The overall training of CAE-LSTM is performed by exploring the reconstruction loss to pursue high-quality reconstruction from topics to region-level features, and the sequence-level reward (e.g., CIDEr) plus coverage reward in self-critical training to encourage the maximum coverage of objects in the paragraph. An overview of our model is depicted in Figure \ref{fig:2}.

\subsection{Notation}

Suppose we have an image ${I}$ to be described by a paragraph $\mathcal{P}$, where $\mathcal{P} = \{s_k\}^K_{k=1}$ consisting of $K$ sentences. Each sentence $s_k = \{w_{(k,t)}\}^{T_k}_{t=1}$ is composed of $T_k$ words and each word $w_{(k,t)}$ is represented as $D_s$-dimensional textual feature $\textbf{w}_{(k,t)}$. Faster R-CNN \cite{ren2015faster} is firstly leveraged to produce a set of detected objects $\mathcal{V}=\{{r_m}\}^{M}_{m=1}$ with $M$ image regions in ${I}$. ${{\bf{v}}^0_m}\in {{\mathbb{R}}^{D_0}}$ denotes the $D_0$-dimensional feature of each image region $r_m$.

\subsection{CAE for Topic Modeling}

The most typical way to distill the topics from an image is to encode region-level features into a global vector via mean pooling and a LSTM-based decoder is subsequently utilized to recursively output topics conditioning on the global vector, while the inherent structure among all regions is not fully exploited. Furthermore, there is no clear evidence that the outputs of LSTM could characterize the topics well. Here we devise a Convolutional Auto-Encoding (CAE) module for topic modeling. The spirit behind follows the philosophy that the holistical topics abstraction over all region-level features via convolutional encoder and the reconstruction from topics to features via deconvolutional decoder can enforce the learnt topics to be more representative and informative.

\paragraph{Convolutional Encoder.} Given the set of detected regions $\mathcal{V}=\{{r_m}\}^{M}_{m=1}$, a convolutional encoder is leveraged to encapsulate all the image regions into $K$ topic vectors, which is purely applied with convolutions. Specifically, each region ${{\bf{v}}^0_m}$ is firstly embedded into a $D_1$-dimensional region-level feature ${{\bf{v}}_m}$ through a linear layer. A region feature map ${\textbf{V}}\in{\mathbb{R}^{M \times D_1 \times 1}}$ is thus constructed by concatenating all region-level features (i.e., ${{\bf{v}}_m}$ is the $m$-th column of ${\textbf{V}}$), which will be set as the input of convolutional encoder. Here $M$, $D_1$, and $1$ denote the width, height, and the number of channels for the input feature map ${\textbf{V}}$, respectively. Next, we utilize a convolutional layer ($conv$) to encode the feature map ${\textbf{V}}$ into the topic feature map $\textbf{V}^s$ consisting of topic vectors:
\begin{equation}
\label{eq:01}
\textbf{V}^s=conv (\textbf{V})\in{\mathbb{R}^{1 \times D_2 \times K}}.
\end{equation}
For the convolutional layer $conv$, the size of the convolutional filter is set as $M \times C_1$, the stride size is $C_2$, and the filter number is $K$. Here each vector along channel dimension within the topic feature map $\textbf{V}^s$ can be regarded as the $k$-th distilled topic vector $\textbf{v}_{k}^s\in{\mathbb{R}^{D_2}}$.

\paragraph{Deconvolutional Decoder.} For deconvolutional decoder, we perform deconvolution (i.e., the conjugate operation of convolution) to decode the distilled topic feature map, $\textbf{V}^s$, back to the region feature map. In particular, given the distilled topic feature map $\textbf{V}^s$, a deconvolutional layer ($deconv$) is adopted to reproduce the region feature map $\tilde{\textbf{V}}$:
\begin{equation}
\label{eq:02}
\tilde{\textbf{V}}=deconv (\textbf{V}^s)\in{\mathbb{R}^{M \times D_1 \times 1}}.
\end{equation}
The filter size and the stride size in $deconv$ are set the same as the convolutional layer $conv$. Here the $m$-th column of the reconstructed region feature map $\tilde{\textbf{V}}$ corresponds to the reconstructed region-level feature for $m$-th image region.

\paragraph{Reconstruction Loss.} To measure the quality of reconstruction for our CAE module, a global reconstruction loss is defined as the L$_1$ distance between the input region feature map $\textbf{V}$ and reconstructed region feature map $\tilde{\textbf{V}}$:
\begin{equation}
\label{eq:03}
L_{rec}(\tilde{\textbf{V}},\textbf{V}) = {\parallel\tilde{\textbf{V}}-\textbf{V} \parallel}_1.
\end{equation}
We also experimented with L$_2$ distance as the reconstruction loss but that did not make a major difference. By minimizing the reconstruction loss, the distilled topic vectors are enforced to capture holistic and representative information from all regions through achieving higher reconstruction quality.

Note that besides acting as the gists to guide the sentence generation in the followed two-level LSTM networks, the distilled topic vectors can also be utilized to determine the number of generated sentences. Specifically, each topic vector is additionally injected into a linear layer to obtain a distribution over two states \{CONTINUE=0, STOP=1\}, which determines whether the sentence is the last one in the paragraph.

\begin{figure}[!tb]
\centering{\includegraphics[width=0.48\textwidth]{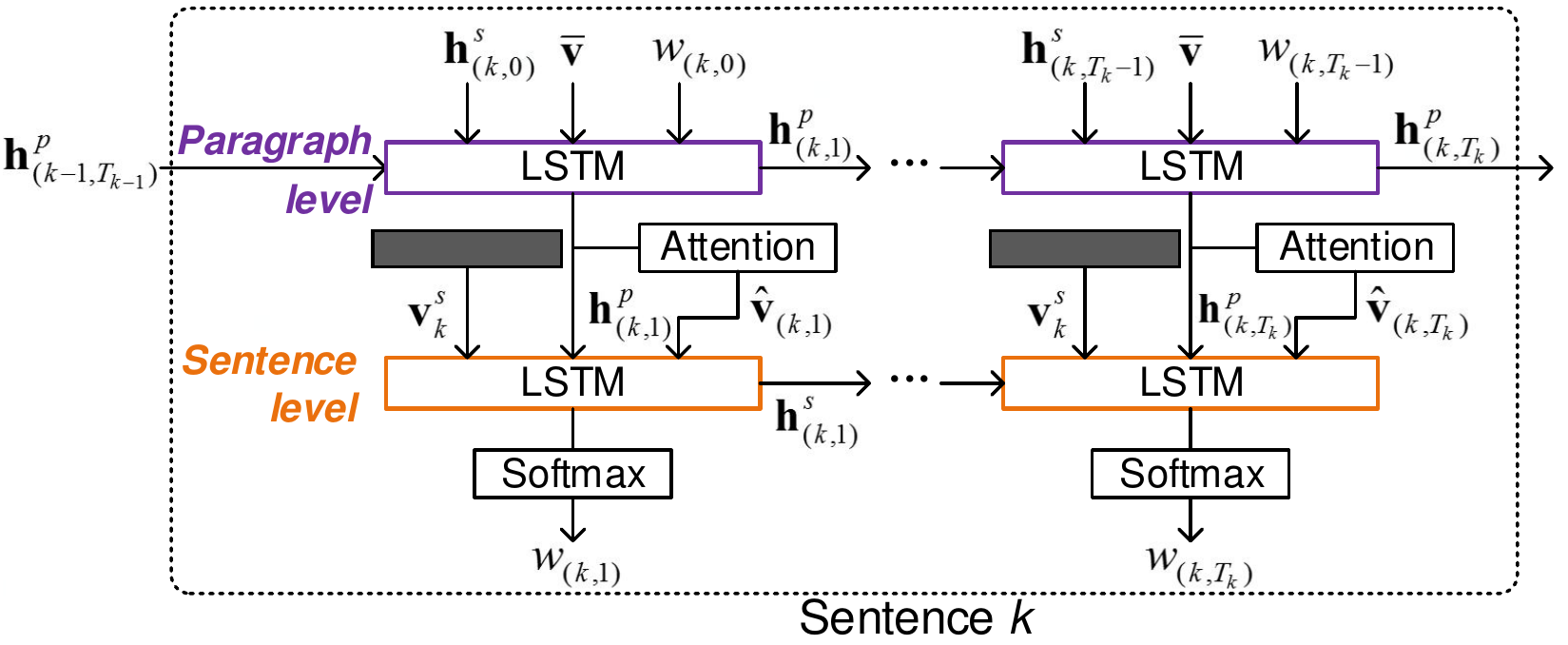}}
\caption{Illustration of two-level LSTM-based paragraph generator.}
\label{fig:3}
\end{figure}

\subsection{Two-level LSTM-based Paragraph Generator}
Inspired by the two-layer LSTM structure with top-down attention mechanism in \cite{anderson2017bottom}, we remould a two-level LSTM-based paragraph generator consisting of a paragraph-level LSTM for inter-sentence dependency modeling and a sentence-level LSTM for sentence generation conditioning on each distilled topic, as illustrated in Figure \ref{fig:3}.

To be specific, at each time step $t$ for generating the $k$-th sentence in a paragraph, the input vector $\textbf{x}_{(k,t)}^p$ of paragraph-level LSTM is defined as the concatenation of the previous output of sentence LSTM $\textbf{h}_{(k,t-1)}^s\in{\mathbb{R}^{H}}$, the mean-pooled image feature $\bar{\textbf{v}} = \frac{1}{M}\sum_{m=1}^M \textbf{v}_m$, and the embedding of previously generated word $\textbf{w}_{(k,t-1)}$:
\begin{equation}\small
\label{eq:1}
\textbf{x}_{(k,t)}^p=[\textbf{h}_{(k,t-1)}^s, \bar{\textbf{v}}, \textbf{W}_s\textbf{w}_{(k,t-1)}],
\end{equation}
where $\textbf{W}_s$ is transformation matrix for input word. Such input collects the maximum contextual information for paragraph-level LSTM to model inter-sentence dependency.

Next, given the output $\textbf{h}_{(k,t)}^p\in{\mathbb{R}^{H}}$ of paragraph-level LSTM and the corresponding topic vector $\textbf{v}_{k}^s\in{\mathbb{R}^{D_2}}$, a normalized attention distribution over all regions is measured as
\begin{equation}\small
\label{eq:2}
\begin{array}{l}
    a^m_{(k,t)} = \textbf{W}_{att} [tanh(\textbf{W}_v\textbf{v}_m+\textbf{W}_h\textbf{h}_{(k,t)}^p+\textbf{W}_t\textbf{v}^s_k)],\\
    \bm{\alpha}_{(k,t)} = softmax(\textbf{a}_{(k,t)}),
\end{array}
\end{equation}
where $a^m_{(k,t)}$ is the $m$-th element of $\textbf{a}_{(k,t)}$, $\textbf{W}_{att}\in{\mathbb{R}^{1 \times D_3}}$, $\textbf{W}_v\in{\mathbb{R}^{D_3 \times D_1}}$, $\textbf{W}_h\in{\mathbb{R}^{D_3 \times H}}$, and $\textbf{W}_t\in{\mathbb{R}^{D_3 \times D_2}}$ are transformation matrices, respectively. Here the $m$-th element ${\alpha}^m_{(k,t)}$ of the attention distribution $\bm{\alpha}_{(k,t)}$ denotes the attention probability of $\textbf{v}_m$. The attended image feature $\hat{\textbf{v}}_{(k,t)} = \sum_{m=1}^{M}\alpha^m_{(k,t)}\textbf{v}_m$ is thus calculated by aggregating all region-level features weighted by attention.

We further set the concatenation of the attended image feature $\hat{\textbf{v}}_{(k,t)}$, the corresponding topic vector $\textbf{v}_{k}^s$ and the output $\textbf{h}_{(k,t)}^p$ of paragraph-level LSTM, as the input $\textbf{x}_{(k,t)}^s$ of sentence-level LSTM for topic-oriented sentence generation:
\begin{equation}\small
\label{eq:5}
\textbf{x}_{(k,t)}^s=[\hat{\textbf{v}}_{(k,t)}, \textbf{v}^s_k, \textbf{h}_{(k,t)}^p].
\end{equation}
The output $\textbf{h}_{(k,t)}^s$ of sentence-level LSTM is fed into a softmax layer to predict the next word $w_{(k,t)}$.

Note that the initial state of sentence-level LSTM is set to zero at every beginning of sentence generation. The~state of paragraph-level LSTM is initialized with zero~only at the beginning of the entire paragraph generation since paragraph-level LSTM should explore inter-sentence dependency recursively throughout whole procedure of paragraph generation.

\subsection{Self-critical Training with Coverage Reward}

To further boost our paragraph generation model by amending the discrepancy between training and inference, we adopt the self-critical training strategy \cite{Rennie2016Self} to optimize two-level LSTM-based paragraph generator. Despite having high quantitative scores, qualitative analysis shows that the captioning models solely optimized with sequence-level loss/reward are often limited to describing very generic information of objects, or rely on prior information and correlations from training examples, and resulting frequently in undesired effects such as object hallucination \cite{lu2018neural}. As a result, besides the sequence-level reward (e.g., CIDEr), we additionally include a coverage reward in self-critical training strategy to encourage the global coverage of objects in the paragraph and thus improve the paragraph.

Specifically, we select the top-1$k$ frequent objects from the vocabulary in the training data as the high-frequency objects. The coverage reward is defined as the coverage ratio of high-frequency objects for the generated paragraph relative to ground-truth paragraph: $R^c =\frac{|Q_g\cap Q_{gt}|}{|Q_{gt}|}$, where $|\cdot|$ denotes the number of elements in a set. Here $Q_g$ and $Q_{gt}$ represent the set of high-frequency objects mentioned in the generated paragraph and ground-truth paragraph respectively. Accordingly, the final reward in self-critical training strategy is measured as the combination of CIDEr reward ($R^d$) and coverage reward: $R = \beta R^c + R^d$, where $\beta$ is the tradeoff parameter.

\section{Experiments}

\subsection{Datasets and Experimental Settings}

\paragraph{Dataset.} We conducted the experiments and evaluated our CAE-LSTM on Stanford image paragraph dataset (Stanford) \cite{krause2017hierarchical}, a benchmark in the field of image paragraph generation. The dataset contains 19,551 images and there is one human-annotated paragraph per image. On average, each paragraph has 67.5 words and each sentence consists of 11.91 words. In our experiments, we follow the widely used settings in \cite{krause2017hierarchical} and take 14,575 images for training, 2,487 for validation and 2,489 for testing.

\paragraph{Compared Methods.} We compare the proposed method with the following state-of-the-art methods: (1) \textbf{Image-Flat} \cite{karpathy2015deep} is a standard image captioning model which directly decodes an image into a paragraph word by word, via a single LSTM. (2) \textbf{Regions-Hierarchical} \cite{krause2017hierarchical} adopts a hierarchical LSTM to generate a paragraph, sentence by sentence. (3) \textbf{RTT-GAN} \cite{liang2017recurrent} integrates sentence attention and word attention into the hierarchical LSTM, coupled with adversarial training strategy. (4) \textbf{CapG-RevG} \cite{chatterjee2018diverse} leverages coherence vectors/global topic vectors to generate coherent paragraphs and maintains the diversity of the paragraphs by a variational auto-encoder formulation. (5) \textbf{CAE-LSTM} is the proposal in this paper. Moreover, one degraded version of CAE-LSTM, i.e., \textbf{LSTM-ATT}, is devised to model topics via LSTM (instead of CAE) and adopt the same two-level LSTM architecture with attention for paragraph generation, which is trained without self-critical training strategy.

\paragraph{Settings.}
For each image, we apply Faster R-CNN to detect objects within this image and select top $M=50$ regions with highest detection confidences to represent the image. Each region is represented as the 4,096-dimensional output of fc7 layer after RoI pooling from the conv5\underline{~}3 feature map of Faster R-CNN (backbone: VGG16 \cite{simonyan2014very}). The Faster R-CNN is pre-trained over Visual Genome \cite{krishna2017visual}, similar to \cite{anderson2017bottom}. To build the vocabulary, all the words from the training set are converted to lowercase and those which appear less than 4 times are omitted. Each word is represented as ``one-hot" vector (binary index vector in a vocabulary). The maximum sentence number $K$ is 6 and the maximum word number in a sentence is 20 (padded where necessary). For our CAE, the convolutional filter size in the convolutional layer is set as $C_1=26$ with stride $C_2=2$. The dimensions of the embedded region-level feature and distilled topic vector are set as $D_1=1,024$ and $D_2=500$. For the two-level LSTM networks, the dimension of hidden state in each LSTM is $H=1,000$. The dimension of the hidden layer for measuring attention distribution is $D_3=512$. Three popular metrics are adopted: METEOR \cite{lavie2014meteor}, CIDEr \cite{vedantam2015cider}, and BLEU-4 \cite{papineni2002bleu}. We compute all metrics with the released code from Microsoft COCO Evaluation Server \cite{chen2015microsoft}.

\paragraph{Implementation Details.}
We follow the two-phrase training recipe in \cite{Rennie2016Self} to train our CAE-LSTM. For the first phrase, we set the learning rate as $1\times10^{-4}$ and the training of CAE-LSTM is thus performed by jointly leveraging the reconstruction loss in CAE and the cross entropy loss in two-level LSTM paragraph generator. Here we evaluate the model at each epoch on the validation set and select the model with the best CIDEr score as an initialization for the next training phrase. For the second phrase of self-critical training, the learning rate is set as $5\times10^{-6}$ and CAE-LSTM is further optimized with the combination of CIDEr reward and coverage reward. During inference, we adopt inference constraint in \cite{melas2018training} to penalize trigram repetition. The tradeoff parameter $\beta$ is set as 8 according to the validation performance. Note that Batch normalization \cite{ioffe2015batch} and dropout \cite{srivastava2014dropout} (dropout rate: $0.5$) are applied in our experiments.

One issue of training our CAE-LSTM is to determine the order of the regions in the concatenation of region-level features as the input of CAE module. In the experiments, the region orders are determined according to the objectiveness score of each image region. We additionally studied the region orders by sorting all the regions with regard to the confidence score of bounding box, or simply using fixed random order. The CIDEr score constantly fluctuates within 0.008 when using different orders of regions for concatenation in our CAE-LSTM. The results indicate that the performance is not sensitive to the selection of region order.

\begin{table}[!tb]\small
\centering
\setlength{\tabcolsep}{3pt}
\scalebox{1}{
\begin{tabular}{l|c c c}
\hline
Method                                             & C & M & B-4 \\ \hline
Image-Flat \cite{karpathy2015deep}            & 11.06  & 12.82   & 7.71   \\
Regions-Hierarchical \cite{krause2017hierarchical}                & 13.52  & 15.95  & 8.69   \\
RTT-GAN \cite{liang2017recurrent}                    & 20.36   & 18.39  & 9.21   \\
CapG-RevG \cite{chatterjee2018diverse} & 20.93   & 18.62   & 9.43   \\ \hline
LSTM-ATT                        & 20.17  & 17.72  & 8.72   \\
CAE-LSTM                           & \textbf{25.15}  & \textbf{18.82}  & \textbf{9.67}   \\\hline
\end{tabular}}
\caption{Performance of our CAE-LSTM and other state-of-the-art methods on Stanford, where C, M, and B-4 are short for CIDEr, METEOR, and BLEU-4. All values are reported as percentage (\%).}
\label{tabel:1}
\end{table}

\subsection{Performance Comparison and Analysis}
\paragraph{Quantitative Analysis.} The performances of different models on the Stanford dataset are shown in Table~\ref{tabel:1}. Overall, the results across three evaluation metrics consistently indicate that our proposed CAE-LSTM achieves better performances against other state-of-the-art techniques including non-attention models (Image-Flat, Regions-Hierarchical, and CapG-RevG) and attention-based approaches (LSTM-ATT and RTT-GAN). Specifically, the CIDEr and METEOR scores of our CAE-LSTM can achieve 25.15\% and 18.82\%, which makes the relative improvement over the best competitor CapG-RevG by 20.2\% and 1.1\%, respectively. As expected, by additionally modeling topics/gists in an image via LSTM, Regions-Hierarchical exhibits better performance than Image-Flat which leaves inter-sentence dependency unexploited. Moreover, LSTM-ATT leads to a performance boost over Regions-Hierarchical, which directly encodes an image as a global representation by performing mean pooling over all region-level features. The results basically indicate the advantage of region-level attention mechanism in the two-level LSTM networks by learning to focus on the image regions that are most indicative to infer the next word. Most specifically, RTT-GAN and CapG-RevG by modeling reality and diversity of paragraphs with Generative Adversarial Networks and Variational Auto-Encoders, improves LSTM-ATT. However, the performances of RTT-GAN and CapG-RevG are still lower than our CAE-LSTM which exploits the inherent structure among all image regions for topic modeling in a convolutional and deconvolutional auto-encoding framework.

\begin{figure*}[!tb]
\centering
\includegraphics[width=1\textwidth]{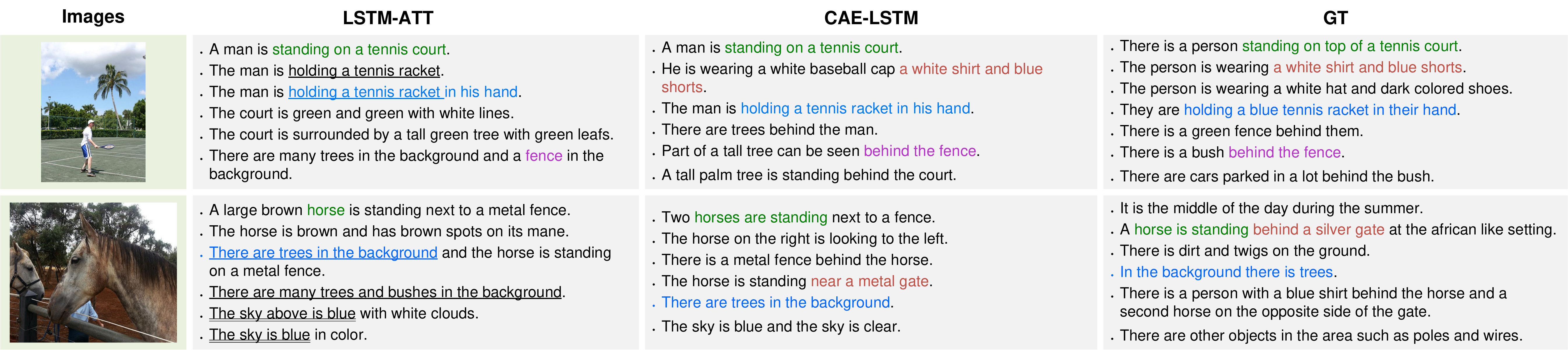}
\caption{Examples for paragraphs generated by LSTM-ATT, CAE-LSTM and human-annotated Ground Truth paragraphs (GT) on Stanford dataset (better viewed in color). The words coded with same colors
indicate the accurate semantic matches between the generated paragraphs and ground truth paragraphs. We underline the semantically similar phrases/sentences in a paragraph.}
\label{fig:4}
\end{figure*}

\paragraph{Qualitative Analysis.}
Figure \ref{fig:4} shows several paragraph examples generated by LSTM-ATT, CAE-LSTM and one human-annotated Ground Truth (GT) paragraph. From these exemplar results, it is easy to see that all of these paragraph generation models can produce somewhat relevant paragraphs, while our proposed CAE-LSTM can generate coherent and accurate paragraphs by learning to distill the gists/topics from an image via a convolutional auto-encoding module to guide paragraph generation. For example, compared to LSTM-ATT which repeatedly generates the phrase ``holding a tennis racket" in the $2$-th and $3$-th sentences for the first image, the produced paragraph for our CAE-LSTM is more coherent to describe the diverse and accurate topics in the image. The results again verify that the convolutional auto-encoding module in our CAE-LSTM could characterize the topics better than LSTM. Moreover, by additionally measuring the coverage of objects as a reward in the self-critical training strategy, our CAE-LSTM is encouraged to produce paragraphs which cover more objects in images, leading to more descriptive paragraph with objects, e.g., ``shirt" and ``shorts."

\begin{figure}[!tb]
\centering
\includegraphics[width=0.5\textwidth]{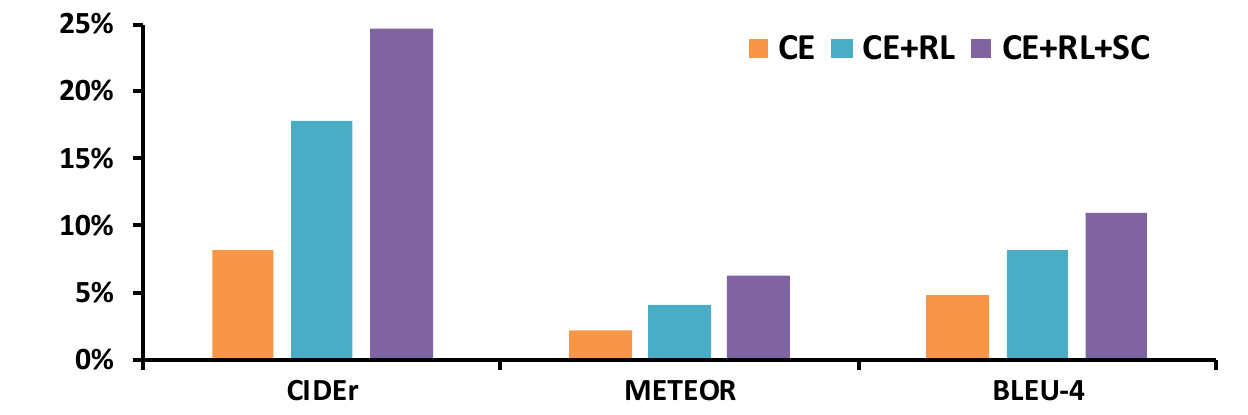}
\caption{The performance improvement by considering one more design (i.e., Convolutional Encoding (\textbf{CE}), Reconstruction Loss (\textbf{RL}), and Self-critical training with Coverage (\textbf{SC})) in CAE-LSTM, over the performance of LSTM-ATT across different metrics.}
\label{fig:5}
\end{figure}

\paragraph{Ablation study.}
Here we study how each design in our CAE-LSTM influences the overall performance. Convolutional Encoding (\textbf{CE}) replaces the LSTM for topic modeling in LSTM-ATT with the convolutional encoder in our devised CAE module. Reconstruction Loss (\textbf{RL}) enforces the distilled topics to capture holistic and representative information via reconstruction from topics to region-level features. Self-critical training with Coverage (\textbf{SC}) further encourages global coverage of objects with a coverage reward in self-critical training. Figure \ref{fig:5} shows the performance improvement by considering one more design for paragraph generation in our CAE-LSTM, over the performance of LSTM-ATT across different metrics. The results across different metrics consistently indicate that the topic modeling via convolutional encoding in CE leads to a performance boost compared to LSTM-ATT which capitalizes on LSTM to explore dependency among topics. Furthermore, by integrating reconstruction loss into the learning of topics, CE+RL exhibits better performance than CE. The results demonstrate the advantage of reconstruction from topics to region-level features in our CAE. By additionally exploiting coverage reward in self-critical training, CE+RL+SC further boosts up the performances.

\paragraph{Human Evaluation.}
To better understand how satisfactory are the paragraphs generated from different methods, we conducted Turing Test to evaluate our CAE-LSTM against the baseline LSTM-ATT. In particular, six evaluators are invited and a subset of 1,000 images is randomly selected from testing set for human evaluation. We show all the evaluators once only one paragraph generated by different approach or human annotation and they are asked: can you determine whether the given paragraph has been generated by a system or by a human being? From evaluators' responses, we calculate the percentage of paragraphs that pass the Turing Test. The results of Turing Test for Human, CAE-LSTM, and LSTM-ATT are 88.5\%, 39.8\%, and 14.7\%, which clearly show that our CAE-LSTM outperforms LSTM-ATT.

\section{Conclusion}
We have present Convolutional Auto-Encoding plus Long Short-Term Memory (CAE-LSTM), which explores the modeling of sentence topics to boost image paragraph generation. Particularly, we study the problem from the viewpoint of topic distillation in an auto-encoding manner. To verify our claim, we have devised a purely convolutional structure, in which an encoder abstracts topics by employing convolutions over region-level image features and a deconvolutional decoder is to validate the topics through high-quality reconstruction from topics to features. A two-level LSTM-based paragraph generator is then remould, on one hand, to model dependency across sentences in a paragraph, and on the other, to generate the sentence conditioning on each learnt topic. Experiments conducted on Stanford image paragraph dataset verify our proposal and analysis. Performance improvements are observed when comparing to state-of-the-art image paragraph generation techniques.

\section*{Acknowledgements}
This work was partially supported by the National Natural Science Foundation of China (Grant No. 61732007 and U1611461).

\bibliographystyle{named}
\bibliography{ijcai19}

\end{document}